\documentclass[a4paper]{nusthesis}
\dsp 
\usepackage{hyperref}
\usepackage[utf8]{inputenc}
\usepackage[english]{babel}
\usepackage{csquotes}

\usepackage{indentfirst} 
\usepackage{silence} 
\usepackage{bookmark}

\usepackage[
  style=ext-authoryear,
  backend=biber,
  maxcitenames=1,
  giveninits=true,
  sorting=nyt,
  dashed=false,
  doi=false]{biblatex}
\DeclareFieldFormat*{title}{``#1''\newunitpunct} 
\newcommand{\citet}[1]{%
	\AtNextCite{\defcounter{maxnames}{1}\defcounter{minnames}{1}}%
	\citeauthor{#1}%
}

\usepackage{microtype} 
\usepackage{tabulary} 
\usepackage{hhline}

\usepackage{enumitem}

\usepackage[noabbrev]{cleveref}
\creflabelformat{equation}{#2\textup{#1}#3}

\usepackage{listings}

\usepackage{amsthm}
\theoremstyle{definition}
 
\theoremstyle{plain}

\usepackage{mathtools}

\usepackage{interval}
\intervalconfig{soft open fences}

\usepackage[linesnumbered,ruled,vlined]{algorithm2e}
\SetAlgorithmName{Algorithm}{Algorithm}{Algorithm}
 
\IncMargin{0.5em}
\SetCommentSty{textnormal}
\SetNlSty{}{}{:}
\SetAlgoNlRelativeSize{0}
\SetKwInput{KwGlobal}{Global}
\SetKwInput{KwPrecondition}{Precondition}
\SetKwProg{Proc}{Procedure}{:}{}
\SetKwProg{Func}{Function}{:}{}
\SetKw{And}{and}
\SetKw{Or}{or}
\SetKw{To}{to}
\SetKw{DownTo}{downto}
\SetKw{Break}{break}
\SetKw{Continue}{continue}
\SetKw{SuchThat}{\textit{s.t.}}
\SetKw{WithRespectTo}{\textit{wrt}}
\SetKw{Iff}{\textit{iff.}}
\SetKw{MaxOf}{\textit{max of}}
\SetKw{MinOf}{\textit{min of}}
\SetKwBlock{Match}{match}{}{}

\usepackage{array}
\newcolumntype{L}[1]{>{\raggedright\let\newline\\\arraybackslash\hspace{0pt}}m{#1}}
\newcolumntype{C}[1]{>{\centering\let\newline\\\arraybackslash\hspace{0pt}}m{#1}}
\newcolumntype{R}[1]{>{\raggedleft\let\newline\\\arraybackslash\hspace{0pt}}m{#1}}

\usepackage{color}
\usepackage[usenames,dvipsnames]{xcolor}
\usepackage{soul}
\soulregister\cite7
\soulregister\ref7
\soulregister\pageref7
\soulregister\autoref7
\soulregister\eqref7

\renewcommand{\eqref}{Equation~\ref}

\newcommand*{\RootPicDir}{pic}
\newcommand*{\PicDir}{\RootPicDir}

\newcommand*{\SetPicSubDir}[1]{\renewcommand*{\PicDir}{\RootPicDir /#1}}
\newcommand*{\Pic}[2]{\PicDir /#2.#1}

\newcommand*{\RootExpDir}{exp}
\newcommand*{\ExpDir}{\RootExpDir}

\newcommand*{\SetExpSubDir}[1]{\renewcommand*{\ExpDir}{\RootExpDir /#1}}

\newcommand*{\BeforeCaptionVSpace}{1ex}

\DeclareOuterCiteDelims{parencite}{\bibopenparen}{\bibcloseparen}

\DeclareFieldFormat{citehyperref}{%
  \DeclareFieldAlias{bibhyperref}{noformat}
  \bibhyperref{#1}}

\DeclareCiteCommand{\cite}
  {\usebibmacro{prenote}}
  {\usebibmacro{citeindex}%
   \printtext[citehyperref]{\usebibmacro{cite}}}
  {\multicitedelim}
  {\usebibmacro{postnote}}

\DeclareCiteCommand*{\cite}
  {\usebibmacro{prenote}}
  {\usebibmacro{citeindex}%
   \printtext[citehyperref]{\usebibmacro{citeyear}}}
  {\multicitedelim}
  {\usebibmacro{postnote}}

\DeclareCiteCommand{\parencite}[\mkouterparencitedelims]
  {\usebibmacro{prenote}}
  {\usebibmacro{citeindex}%
   \printtext[citehyperref]{\usebibmacro{cite}}}
  {\multicitedelim}
  {\usebibmacro{postnote}}

\DeclareCiteCommand{\footcite}[\mkbibfootnote]
  {\usebibmacro{prenote}}
  {\usebibmacro{citeindex}%
  \printtext[citehyperref]{ \usebibmacro{cite}}}
  {\multicitedelim}
  {\usebibmacro{postnote}}

\DeclareCiteCommand{\footcitetext}[\mkbibfootnotetext]
  {\usebibmacro{prenote}}
  {\usebibmacro{citeindex}%
   \printtext[citehyperref]{\usebibmacro{cite}}}
  {\multicitedelim}
  {\usebibmacro{postnote}}

\DeclareCiteCommand{\textcite}
  {\boolfalse{cbx:parens}}
  {\usebibmacro{citeindex}%
   \printtext[citehyperref]{\usebibmacro{textcite}}}
  {\ifbool{cbx:parens}
     {\bibcloseparen\global\boolfalse{cbx:parens}}
     {}%
   \multicitedelim}
  {\usebibmacro{textcite:postnote}}
  
\begin{document}

\title{Planning using Schrödinger Bridge Diffusion Models}

\author{Adarsh Srivastava}
\prevdegrees{%
  B.E.(Hons.), BITS Pilani}
\degree{Master of Computing}
\field{Artificial Intelligence}
\degreeyear{2024}
\supervisor{Assistant Professor Harold Soh}

\examiners{%
  Assistant Professor Bryan Hooi
  }

\maketitle

\declaredate{23 May 2024}
\declaresign{\RootPicDir /signature.png} 
\declarationpage

\begin{frontmatter}
  \begin{acknowledgments}

I would like to thank Prof. Harold Soh for giving me this opportunity, and for his time, guidance and patience with me throughout the thesis. 

I am also grateful to Alvin Heng, a Ph.D. student under Prof. Soh, with whom I had many fruitful discussions and who patiently walked me through several difficult concepts, and whose discussions with Prof. Soh helped spark many ideas, including the one for this thesis. Simply listening to his discussions has allowed me to learn much and think in newer directions.

I am also grateful to my friends in NUS and beyond, whose constant encouragement kept me going. 

\end{acknowledgments}

  \tableofcontents 
  \begin{abstract}

Offline planning often struggles with poor sampling efficiency as it tries to learn policies from scratch. Especially with diffusion models, such \textit{cold start} practices mean that both training and sampling become very expensive. We hypothesize that certain environment constraint priors or cheaply available policies make it unnecessary to learn from scratch, and explore a way to incorporate such priors in the learning process. To achieve that, we borrow a variation of Schrödinger bridge formulation from image-to-image setting and apply it to planning tasks. We study the performance on some planning tasks and compare the performance against the DDPM formulation. 

\end{abstract}

\end{frontmatter}

\SetPicSubDir{1-intro}

\chapter{Introduction}
\vspace{2em}

Planning from scratch has long struggled with poor sample efficiency and therefore the need for a large amount of data and compute. Learning from scratch is not only expensive, but also sometimes unnecessary, either because natural domain or task constraints already exist, or because an approximate policy is cheaply available. Not being able to incorporate these into the planning algorithm naturally not only leads to higher data requirements ( or poor \textit{sample efficiency}) but also longer training and sampling times, since the model needs to learn information that is readily available and can be supplied as a prior. 

Prior methods that have tried to deal with this issue have either tried to incorporate learned skills as priors between tasks~\parencite{pertsch2020spirl}, using heuristics to accelerate the algorithm~\parencite{hurl}, or warm starting policies~\parencite{pmlr-v202-uchendu23a}.

In this thesis, we explore a method of planning that directly incorporates a prior policy into the training, such that learning starts from the prior policy distribution itself. To do so, we build upon two prior works - using diffusion models to learn trajectory planning and optimization as first described in ~\cite{janner2022diffuser}, and a bridge diffusion model that transports one distribution to another, formulated for image-to-image tasks such as inpainting and deblurring in ~\cite{i2sb}.

Diffusion models~\parencite{sohl-dickstein15}~\parencite{DDPM}~\parencite{song2021denoising} have shown state-of-the-art results in many image and video generation tasks including unconditioned generation, text-conditioned generation, image inpainting, superresolution, and Image-to-Image Translation tasks~\parencite{parmar2023zeroshot}. In ~\cite{janner2022diffuser}, they have also shown their ability to capture complex target distributions and ushered in a new paradigm of learned planning - one where planning and trajectory optimization were wrapped into a single learned model. However, while diffusion models beat other generative models in high-quality sample generation capabilities with mode coverage and diversity as well as stable training, they also typically suffer from expensive sampling and long training times.

\begin{figure}[ht!]
  \centering
  \includegraphics[width=0.4\linewidth]{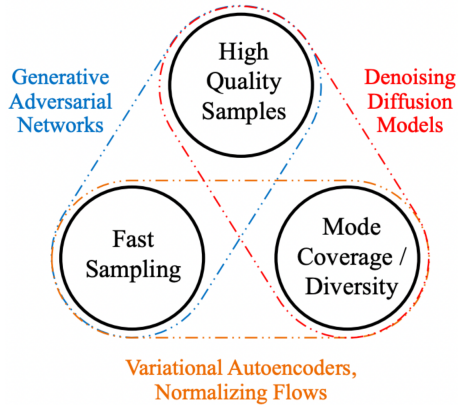}
  \vspace{\BeforeCaptionVSpace}
  \caption{Diffusion models suffer from expensive sampling and long training times.}
  \label{into:fig:ddpm_weakness}
\end{figure}

A key reason for this inefficiency is that diffusion models learn to denoise to the target distribution from pure Gaussian noise. Intuitively, if it was possible to sample from a distribution that was closer to the target distribution than pure noise (in other words, it had some structural information about the target), it should make sense that a similar process as diffusion would have less to learn ( and therefore learn better from fewer samples, i.e., improved sample efficiency), and would be less expensive to sample from (since the generation has less to predict). 

This idea was explored by ~\cite{i2sb} in the context of image-to-image translation tasks using diffusion bridges built upon the theoretical framework of the Schrödinger Bridge problem, but so far has not been applied to planning or imitation learning tasks. We hypothesize that using a bridging method to generate plans should also work, and should lead to better sample efficiency and faster sampling.

Since I\textsuperscript{2}SB~\parencite{i2sb} and Diffuser~\parencite{janner2022diffuser} share the common framework of diffusion models, and I\textsuperscript{2}SB shows good performance on image translation tasks, we pick I\textsuperscript{2}SB as the choice of bridging model to explore how effective a pairing of the two approaches can be in a planning setting. 

This thesis is structured as follows: we first go through the necessary background on diffusion models, the Diffuser approach, the Schrödinger Bridge problem, the I\textsuperscript{2}SB version of Schrödinger Bridges, and their adaptation to image-to-image tasks. We then outline our own approach and lay out our experimental setup. Finally, we discuss the results and future directions.

We show that while the tractable class of Schrödinger Bridge we use to bridge distributions works, and shows better performance than DDPM at very low NFEs, at higher NFEs DDPM catches up and is able to learn and generalize better. We also show how we might construct priors for tasks in planning for such bridging methods. However, this work does not go into the theoretical analysis of the various bridging models, as that is beyond the scope of this work. 

\SetPicSubDir{2-background}
\chapter{Background \& Literature Review}

\label{ch:review}
\vspace{2em}

This chapter outlines the theoretical foundations which this thesis builds upon. First, we give a brief overview of diffusion models in two forms - the probabilistic model~\parencite{DDPM}, and the score-based model~\parencite{song2019}. We also briefly cover the use of diffusion models in planning, and in particular, the Diffuser approach~\parencite{janner2022diffuser}, which we use as the foundation and baseline for our approach. We then cover the theoretical foundations at the crux of our approach - the Schrödinger Bridge, and one particular formulation of it - I\textsuperscript{2}SB ~\parencite{i2sb}, which is the formulation we use in our experiments.

\section{Diffusion models}

Diffusion models are a class of deep generative models inspired by non-equilibrium thermodynamics that iteratively corrupt data with increasing noise, and learn to reverse this process from a noise sample to form a generative model for the data distribution. \textit{Denoising diffusion probabilistic modeling} (DDPM)~\parencite{sohl-dickstein15}~\parencite{DDPM} learns a denoising Markov chain process with discrete time steps, while score-based models~\parencite{song2019} learn the gradient of the log data distribution, and use it to guide a random towards regions of higher data density. In particular, score-based modeling through SDEs~\parencite{song2021scorebased} models the diffusion process through continuous-time forward and reverse SDEs.

\subsection{Diffusion Probabilistic models}

A T-step Denoising Diffusion Probabilistic Model(DDPM) consists of two processes: the forward process (also referred to as the diffusion process), and the reverse inference process.

\begin{figure}[ht!]
  \centering
  \includegraphics[width=.8\linewidth]{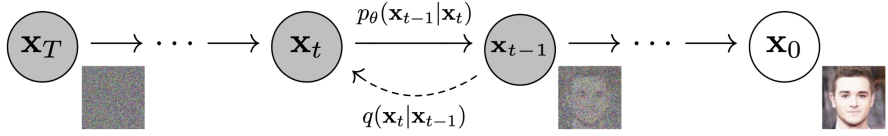}
  \vspace{\BeforeCaptionVSpace}
  \caption{The DDPM graphical model.}
  \label{review:fig:ddpm}
\end{figure}

The forward process, or the diffusion process, adds Gaussian noise to the sample according to a fixed variance schedule \(\beta_{1}, \dotsc , \beta_{T}\):

\begin{equation}
q\left(\mathbf{x}_{1: T} \mid \mathbf{x}_0\right):=\prod_{t=1}^T q\left(\mathbf{x}_t \mid \mathbf{x}_{t-1}\right), \quad q\left(\mathbf{x}_t \mid \mathbf{x}_{t-1}\right):=\mathcal{N}\left(\mathbf{x}_t ; \sqrt{1-\beta_t} \mathbf{x}_{t-1}, \beta_t \mathbf{I}\right)
\end{equation}

There is also a closed-form expression for sampling the forward process at any \(x_t\); using $\alpha_t:=1-\beta_t$ and $\bar{\alpha}_t:=\prod_{s=1}^t \alpha_s$, we have: 

\begin{equation}
q\left(\mathbf{x}_t \mid \mathbf{x}_0\right)=\mathcal{N}\left(\mathbf{x}_t ; \sqrt{\bar{\alpha}_t} \mathbf{x}_0,\left(1-\bar{\alpha}_t\right) \mathbf{I}\right) \label{eq:2.2}
\end{equation}

The reverse process, or the denoising process, is defined as a Markov chain with learned Gaussian transitions starting at $p\left(\mathbf{x}_T\right)=\mathcal{N}\left(\mathbf{x}_T ; \mathbf{0}, \mathbf{I}\right)$:

\begin{equation}
p_\theta\left(\mathbf{x}_{0: T}\right):=p\left(\mathbf{x}_T\right) \prod_{t=1}^T p_\theta\left(\mathbf{x}_{t-1} \mid \mathbf{x}_t\right), \quad p_\theta\left(\mathbf{x}_{t-1} \mid \mathbf{x}_t\right):=\mathcal{N}\left(\mathbf{x}_{t-1} ; \boldsymbol{\mu}_\theta\left(\mathbf{x}_t, t\right), \boldsymbol{\Sigma}_\theta\left(\mathbf{x}_t, t\right)\right) \label{eq:2.3}
\end{equation}

The model learns by training a time-dependent UNet-like network to predict the total noise, $\boldsymbol{\epsilon}_\theta$, added to the sample during the forward process at any timestep \(t\):

\begin{equation}
L_{\text {simple }}(\theta):=\mathbb{E}_{t, \mathbf{x}_0, \boldsymbol{\epsilon}}\left[\left\|\boldsymbol{\epsilon}-\boldsymbol{\epsilon}_\theta\left(\sqrt{\bar{\alpha}_t} \mathbf{x}_0+\sqrt{1-\bar{\alpha}_t} \boldsymbol{\epsilon}, t\right)\right\|^2\right]
\end{equation}

\subsection{Score-Based Generative Modeling through SDEs}

Another approach~\parencite{song2021scorebased} treats the forward process as a continuous time-dependent stochastic differential equation (SDE) given by ($\mathbf{w}$ is the standard Wiener process or Brownian motion):

\begin{equation}
\mathrm{d} \mathbf{x}=\mathbf{f}(\mathbf{x}, t) \mathrm{d} t+g(t) \mathrm{d} \mathbf{w} \label{eq:2.5}
\end{equation}

and the reverse process reverse-time SDE by:

\begin{equation}
\mathrm{d} \mathbf{x}=\left[\mathbf{f}(\mathbf{x}, t)-g(t)^2 \nabla_{\mathbf{x}} \log p_t(\mathbf{x})\right] \mathrm{d} t+g(t) \mathrm{d} \overline{\mathbf{w}} \label{eq:2.6}
\end{equation}

\begin{figure}[ht!]
  \centering
  \includegraphics[width=0.93\linewidth]{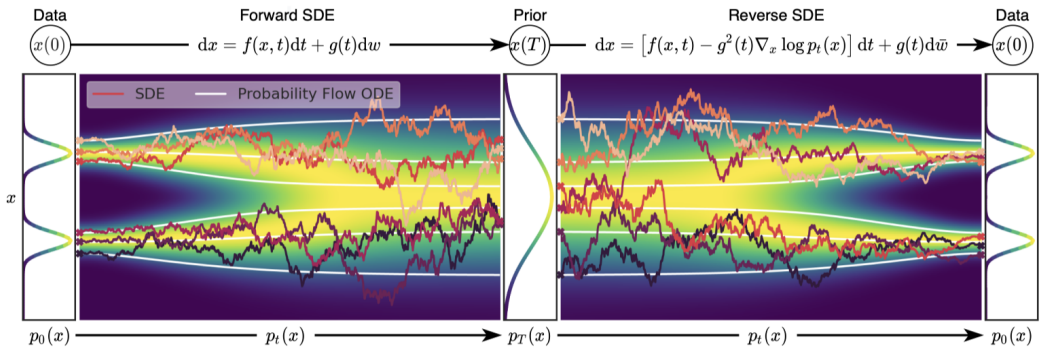}
  \caption{Overview of score-based generative modeling through SDEs.}
  \label{review:fig:score_based_sde}
\end{figure}

The model learns by training a time-dependent UNet-like network to predict the score \(\nabla_{\mathbf{x}} \log p_t(\mathbf{x})\) of the data distribution for the marginal distribution \(p(x)\) and continuous time \(t\). With this score available, sampling is done by integrating the reverse SDE using numerical SDE solvers.

\cite{song2021scorebased} also talk about two kinds of SDEs - the Variance Exploding (VE) SDE, and the Variance Preserving (VP) SDE, depending on whether the variance of the forward SDe explodes as $t \rightarrow \infty$ or remain bounded. In particular, they note that SMLD~\parencite{song2019} is a VE model, and DDPM~\parencite{DDPM} is a VP model.

\newpage

\section{Diffusion models in Planning}

Diffusion models trained to generate images or videos have previously been used in the planning domain in several works. DALL-E-Bot~\parencite{dall-e-bot} uses the DALL-E~\parencite{dall-e} foundation model trained on web-scale data to generate text-conditioned goal images for object rearrangement in a scene. UniPi~\parencite{unipi} uses text-to-video diffusion models to construct a video plan of how to achieve a task, followed by an inverse dynamics model to extract control actions. ROSIE~\parencite{rosie} uses conditional inpainting of the current scene using text-to-image diffusion models to create variations and improve generalization during training without needing more real-world data.

However, one of the earliest works that directly trained diffusion models to capture the underlying trajectory/policy distribution was Diffuser~\parencite{janner2022diffuser} - which used diffusion over state-action pairs to generate viable variable horizon trajectories. This work showed that diffusion models can generate optimized trajectories non-autoregressively over long horizons even for complex tasks, without suffering from compounding errors, and can also compose in-distribution trajectories to generate novel trajectories. Since this thesis builds upon the Diffuser approach, we will explore the technical details of this approach in depth in the following section.

\section{The Diffuser approach}

Traditionally, planning was accomplished by learning a dynamics model of the environment and then using trajectory optimizers to generate the trajectory. The key idea behind the Diffuser~\parencite{janner2022diffuser} approach was to wrap the two into a single diffusion model that learns to generate optimized trajectories for the environment and given constraints. 

\subsection{Architecture}
To reap the advantages of work done on image-based diffusion models, Diffuser treats trajectories it trains on much like an image - the state and action values are analogous to the pixel values, the timesteps being the \textit{length}, and the \textit{width} being the sum of dimensionalities of the state and action values. Specifically, the input (and output) trajectories $\tau$  are defined as the following array, $T$ being the prediction horizon.

$$
\boldsymbol{\tau}=\left[\begin{array}{llll}
\mathbf{s}_0 & \mathbf{s}_1 & \ldots & \mathbf{s}_T \\
\mathbf{a}_0 & \mathbf{a}_1 & \ldots & \mathbf{a}_T
\end{array}\right]
$$

Because planning and trajectory optimization are wrapped into one, and trajectory optimization is inherently non-markovian (the present state depends on both the past and future states), Diffuser predicts (or denoises) all timesteps at once, instead of predicting autoregressively. 

Despite the non-autoregressive prediction, Diffuser maintains temporal local consistency by using a small local receptive field along the planning horizon, implemented using 1D convolutions over the time dimension. Also, because Diffuser uses these temporal convolutions, the architecture is independent of the planning horizon to allow variable-length plans.

\begin{figure}[ht!]
  \centering
  \includegraphics[width=.55\linewidth]{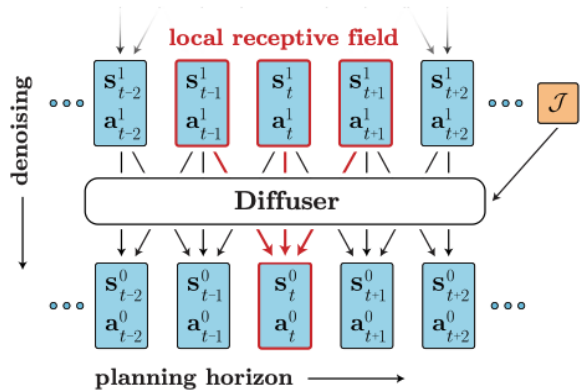}
  \vspace{\BeforeCaptionVSpace}
  \caption{Diffuser uses a local receptive field to enforce local consistency while predicting non-autoregressively.}
  \label{review:fig:diffuser_network}
\end{figure}

Diffuser also allows for a guidance function $\mathcal{J}$ to influence the generated trajectory. This guidance function can represent prior knowledge, rewards, or costs associated with the trajectory. 

Because Diffuser uses a temporal local receptive field to enforce local consistency, an interesting way to generate a goal-seeking policy is to treat the problem as an inpainting problem, where the start and end states are kept fixed, and the model denoises the remaining trajectory. This is the method used in the Maze2D benchmark experiments run by the authors.

\begin{figure}[ht!]
  \centering
  \includegraphics[width=.85\linewidth]{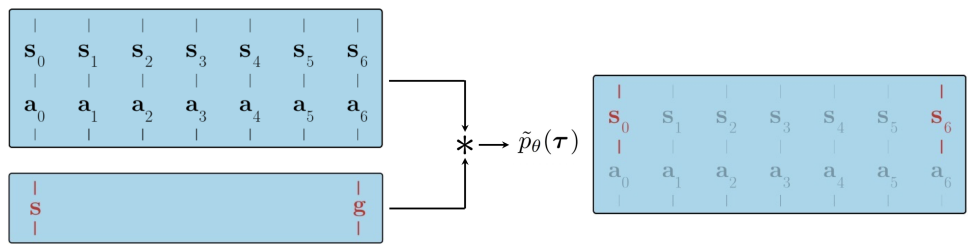}
  \vspace{\BeforeCaptionVSpace}
  \caption{Diffuser treats generating a goal-seeking policy as an inpainting problem.}
  \label{review:fig:goal-seeking-policy}
\end{figure}

The overall architecture of Diffuser remains remarkably similar to image diffusion models, except that it replaces the spatial convolution UNet with a temporal convolution UNet, with 1D convolutions over the planning horizon.

\subsection{Key Results}
The authors show the following key aspects of planning with Diffuser:
\begin{itemize}
  \item Ability to perform \textbf{long horizon planning}, because Diffuser does not suffer from compounding errors, and losses are calculated over the whole trajectory instead of minimizing single step error.
  \item Ability to generate \textbf{variable length plans}, since the architecture is independent of the planning horizon. 
  \item Ability to \textbf{compose in-distribution trajectories} to generate novel trajectories.
\end{itemize}

\begin{figure}[ht!]
  \centering
  \includegraphics[width=.55\linewidth]{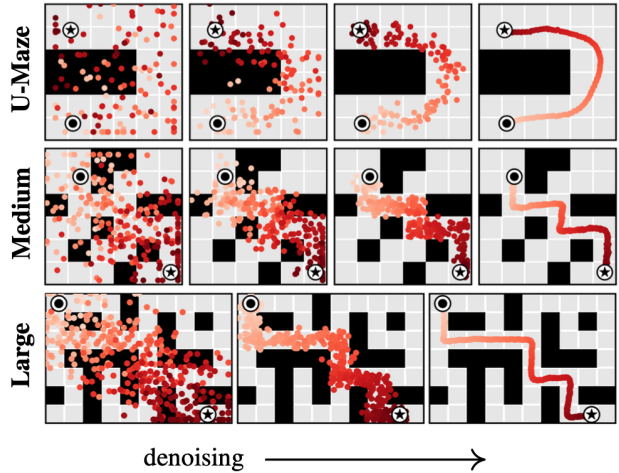}
  \vspace{\BeforeCaptionVSpace}
  \caption{A visualization of non-autoregressive denoising of trajectories on Maze2D tasks.}
  \label{review:fig:inpainting-visualization}
\end{figure}

Diffuser showed competitive results on D4RL benchmarks \textit{across tasks} compared to other state-of-the-art algorithms, once again confirming the ability of Diffusion models to capture complex distributions with relative ease. A limitation of Diffuser is that \textit{individual plans are slow to generate} due to the iterative nature of the algorithm.

\section{Schr\"odinger Bridges}

Directly transporting between two distributions is an important problem, especially in image-to-image translation tasks. A flurry of recent work has happened on coming up with ways to achieve this transport. ~\cite{stochastic-interpolants} introduces Stochastic Interpolants that bridge two arbitrary densities exactly and in finite time, which can be used for image generation and translation. ~\cite{ddbm} propose a unified diffusion bridge method as a natural extension of DDPM~\parencite{DDPM}. 

For this thesis, we borrow the formulation of a tractable class of Schrödinger Bridge from the I\textsuperscript{2}SB paper\parencite{i2sb}, which used the technique on image-to-image translation tasks such as inpainting, super-resolution, and deblurring. In the following sub-sections, we go through the details of their methods, and also lay down some theoretical background. 

\subsection{Schr\"odinger Bridge Problem}
The Schr\"odinger Bridge Problem~\parencite{sbp_1932} is finding the most likely evolution of a stochastic process between two continuous distributions, and is a classical problem appearing in areas of optimal control and probability. It can also be seen as an entropy-regularized optimal transport problem. 

Schr\"odinger Bridges (SB) can also be seen as the general case of a score-based generative model (SGM), since while SGM degrades to and denoises from pure noise, SB goes \textit{beyond} pure noise to another distribution. 

Despite the mathematical generalization, SGM and SB have evolved independently and thus have different computational frameworks to solve them. Schrödinger Bridge models often use iterative projection methods \parencite{chen2023likelihood} which have intractable complexity for high dimensional problems, and the methods are different from how diffusion models are trained.

\begin{figure}[ht!]
  \centering
  \includegraphics[width=0.85\linewidth]{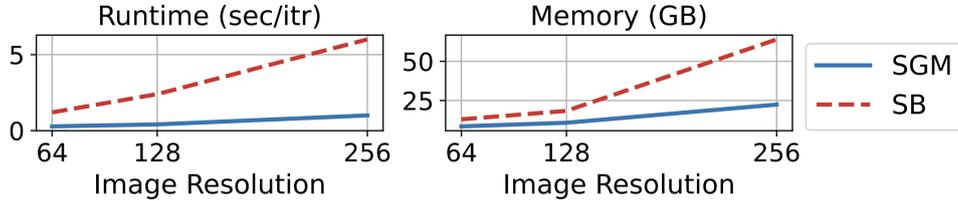}
  \vspace{\BeforeCaptionVSpace}
  \caption{SB is 6x slower and 3x more memory intensive on high dimensional inputs on similar tasks as SGM. \parencite{i2sb}}
  \label{review:fig:sgm_v_sb}
\end{figure}

\subsection{I\textsuperscript{2}SB: Image-to-Image Schrödinger Bridge}
Diffusion bridges have been recently explored in ~\cite{debortoli2023diffusion}, ~\cite{chen2023likelihood}, \cite{i2sb}, and \cite{ddbm}, and have shown to perform well on tasks like image-to-image translation, image inpainting, image restoration, and super-resolution. 

Since we use the same formulation of Schrödinger Bridge as the I2SB paper ~\parencite{i2sb}, we describe below their approach and their theoretical and experimental results.

The key idea behind the I\textsuperscript{2}SB paper was that degraded images still contain a lot of useful information about the final image, and thus diffusing to and from Gaussian white noise, which has no structural information of the target distribution, is bound to take longer to learn, and more diffusion steps to infer. To better leverage the structure of the problem, I\textsuperscript{2}SB directly starts the generative process from the degraded image and builds diffusion bridges between the two distributions. To construct such diffusion bridges, the authors came up with a tractable class of Schrödinger Bridge that works for this problem.

\begin{figure}[ht!]
  \centering
  \includegraphics[width=.6\linewidth]{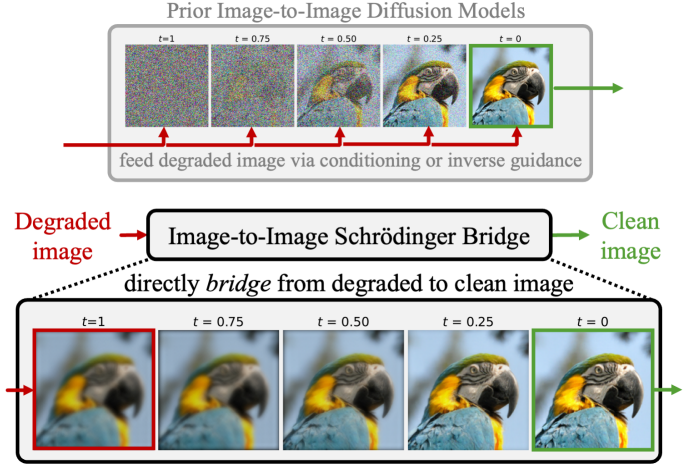}
  \vspace{\BeforeCaptionVSpace}
  \caption{I\textsuperscript{2}SB found a tractable way to directly bridge from degraded to clean images without going to and from Gaussian noise.}
  \label{review:fig:i2sb_idea}
\end{figure}

In the following subsections, we discuss the theory behind the I\textsuperscript{2}SB model, and some of their results on image-to-image tasks.

\subsubsection{Original Schrödinger Bridge SDEs}

Given two boundary distributions $p_{\mathcal{A}}$ and $p_{\mathcal{B}}$ in two distinct domains, with  $X_0 \sim p_{\mathcal{A}}$ and $X_1 \sim p_{\mathcal{B}}$, where $X_t$ is indexed by $t \in[0,1]$, the original Schrödinger Bridge \parencite{sbp_1932} model consider the following forward and backward SDEs:

\begin{equation}
\mathrm{d} X_t=\left[f_t+\beta_t \nabla \log \Psi\left(X_t, t\right)\right] \mathrm{d} t+\sqrt{\beta_t} \mathrm{~d} W_t \label{eq:2.8}
\end{equation}

\begin{equation}
\mathrm{d} X_t=\left[f_t-\beta_t \nabla \log \widehat{\Psi}\left(X_t, t\right)\right] \mathrm{d} t+\sqrt{\beta_t} \mathrm{~d} \bar{W}_t
\end{equation}

The functions $\Psi, \widehat{\Psi} \in C^{2,1}\left(\mathbb{R}^d,[0,1]\right)$ are time-varying energy potentials that solve the following coupled PDEs,

\begin{equation}
\left\{\begin{array}{l}
\frac{\partial \Psi(x, t)}{\partial t}=-\nabla \Psi^{\top} f-\frac{1}{2} \beta \Delta \Psi \\
\frac{\partial \widehat{\Psi}(x, t)}{\partial t}=-\nabla \cdot(\widehat{\Psi} f)+\frac{1}{2} \beta \Delta \widehat{\Psi}
\end{array}\right.
\end{equation}

\begin{equation}
\text { s.t. } \Psi(x, 0) \widehat{\Psi}(x, 0)=p_{\mathcal{A}}(x), \Psi(x, 1) \widehat{\Psi}(x, 1)=p_{\mathcal{B}}(x) \label{eq:2.11}
\end{equation}

It is this coupling in the above equations that gives general SB methods unfavorable complexity on high dimensional tasks compared to SGM. (\Cref{review:fig:sgm_v_sb})

From \eqref{eq:2.5} and \eqref{eq:2.8}, we see the equations differ only in the additional nonlinear drift terms  $\nabla \log \Psi$. The forward drift $\nabla \log \Psi$ is what allows the process to transport samples beyond Gaussian priors. This also shows how SGMs are a special case of general SB.

\subsubsection{ From SB to I\textsuperscript{2}SB}

To go from the general Schrödinger Bridge to a tractable version, the authors make some modifications to the formulation. First, we observe that the nonlinear drifts in \eqref{eq:2.8} resemble the score function in \eqref{eq:2.6} when we view $\Psi(\cdot, t)$ and $\widehat{\Psi}(\cdot, t)$ as densities. Then, we note that $\nabla \log \widehat{\Psi}\left(X_t, t\right)$ and $\nabla \log \Psi\left(X_t, t\right)$ are the score functions of the following linear SDEs respectively:

\begin{equation}
\mathrm{d} X_t=f_t\left(X_t\right) \mathrm{d} t+\sqrt{\beta_t} \mathrm{~d} W_t, \quad X_0 \sim \widehat{\Psi}(\cdot, 0) \label{eq:2.12}
\end{equation}

\begin{equation}
\mathrm{d} X_t=f_t\left(X_t\right) \mathrm{d} t+\sqrt{\beta_t} \mathrm{~d} \bar{W}_t, \quad X_1 \sim \Psi(\cdot, 1)
\end{equation}

Remember that we use neural nets to parameterize the score functions of linear SDE in SGM (\eqref{eq:2.5}). Therefore, if we can parameterize $\nabla \log \widehat{\Psi}$ with the score network, we can use SGM to learn $\nabla \log \widehat{\Psi}$. A similar logic goes for the second equation. 

However, we still cannot sample from $\widehat{\Psi}(\cdot, 0)$ or $\Psi(\cdot, 1)$ because of the coupling in \eqref{eq:2.11}. So the authors present a tractable case with a boundary modification. 

If we let $p_{\mathcal{A}}(\cdot):=\delta_a(\cdot)$ be the Dirac Delta distribution centered at some data point a, we have:

\begin{equation}
\widehat{\Psi}(\cdot, 0)=\delta_a(\cdot), \quad \Psi(\cdot, 1)=\frac{p_{\mathcal{B}}}{\widehat{\Psi}(\cdot, 1)}
\label{eq:2.13}
\end{equation}

This is equivalent to the backward drift driving the reverse process of \eqref{eq:2.12} always flows towards $a$, irrespective of $p_{\mathcal{B}}$. Although this might mean that, in theory, we would not generalize beyond such data points in the training sample, the authors rely on the strong generalization abilities of neural networks to overcome this in practice. 

\subsubsection{Implementation}

Based on the above modifications, the authors come up with a training and sampling plan. Let $X_0$ be the target data point from  $p_{\mathcal{A}}\left(X_0\right)$ and $X_1$ be its degraded pair from  $p_{\mathcal{B}}\left(X_1 \mid X_0\right)$.

\textbf{Training}: The authors derive the forward process posterior of \eqref{eq:2.8} to be the following (derivation is deferred to the original paper):

\begin{equation}
\begin{gathered}
q\left(X_t \mid X_0, X_1\right)=\mathcal{N}\left(X_t ; \mu_t\left(X_0, X_1\right), \Sigma_t\right) \\
\mu_t=\frac{\bar{\sigma}_t^2}{\bar{\sigma}_t^2+\sigma_t^2} X_0+\frac{\sigma_t^2}{\bar{\sigma}_t^2+\sigma_t^2} X_1, \quad \Sigma_t=\frac{\sigma_t^2 \bar{\sigma}_t^2}{\bar{\sigma}_t^2+\sigma_t^2} \cdot I
\label{eq:training}
\end{gathered}
\end{equation}

where $\sigma_t^2:=\int_0^t \beta_\tau \mathrm{d} \tau$ and $\bar{\sigma}_t^2:=\int_t^1 \beta_\tau \mathrm{d} \tau$ are variances accumulated from either sides.

This is analogous in meaning and form to the closed form posterior during training of DDPM in \eqref{eq:2.2}.

\textbf{Sampling}: The sampling procedure for I\textsuperscript{2}SB can be done exactly the same as that of DDPM. However, the authors also show that the same formula used above during training can also be used during sampling. 

During sampling, typically, DDPM needs to recursively run each timestep in \eqref{eq:2.3} one by one with the model generating the required $\boldsymbol{\epsilon}_\theta$. However, in the I\textsuperscript{2}SB formulation, the above posterior also marginalizes the recursive sampling (the authors prove this using induction utilizing \eqref{eq:training}, proof is deferred to the original paper) such that:

\begin{equation}
q\left(X_n \mid X_0, X_N\right)=\int \Pi_{k=n}^{N-1} p\left(X_k \mid X_0, X_{k+1}\right) \mathrm{d} X_{k+1}
\label{closedformsampling}
\end{equation}

Therefore, we can sample multiple diffusion steps in one function call (1 NFE) of the network, though this is not necessary. 

\subsubsection{Key Results}

I\textsuperscript{2}SB uses Frechet Inception Distance (FID) scores and classifier accuracy of a pre-trained ResNet50 to compare against other image-to-image translation algorithms, on 4 primary tasks - 4 x super-resolution, Inpainting, JPEG restoration, and deblurring.

The key takeaway from those results is that I\textsuperscript{2}SB surpasses or matches the performance of the other algorithms, with a much greater sampling efficiency, i.e., with far fewer function calls, which makes sense, since I\textsuperscript{2}SB starts generation from a much more structurally informative prior.

\begin{figure}[ht!]
  \centering
  \includegraphics[width=.5\linewidth]{\Pic{png}{i2sb_result_qual}}
  \vspace{\BeforeCaptionVSpace}
  \caption{I\textsuperscript{2}SB performs better than Palette \parencite{pallete} at lower NFEs, suggesting a higher sampling efficiency.}
  \label{review:fig:i2sb_result_qual}
\end{figure}

\begin{figure}[ht!]
  \centering
  \includegraphics[width=.6\linewidth]{\Pic{png}{i2sb_result_quant}}
  \vspace{\BeforeCaptionVSpace}
  \caption{I\textsuperscript{2}SB has much better FID and CA scores at lower NFEs compared to Palette \parencite{pallete}.}
  \label{review:fig:i2sb_result_quant}
\end{figure}
\SetPicSubDir{3-approach}
\chapter{Methodology}
\label{ch:approach}

\vspace{2em}

To test our hypothesis, we keep the same UNet network, architecture, and structure as the Diffuser approach for planning with DDPM. We then replace the DDPM training and sampling steps with the I\textsuperscript{2}SB Schr\"odinger bridge steps using \Cref{eq:training} and \Cref{closedformsampling} respectively.

There is an additional component to our setup - the prior. Whereas I\textsuperscript{2}SB used the degraded version (such as blurred, or masked) of the target image as the originating boundary pair, we are free to design our own prior. Before we do that, we outline below the conditions such a prior must satisfy to be an effective candidate, based on the theoretical discussion in the preceding sections:

\begin{itemize}
 \item Let $X_0$ be the target trjectory from  $p_{\mathcal{A}}\left(X_0\right)$ and $X_1$ be the prior. Then,  $X_1$ must be sampled from  $p_{\mathcal{B}}\left(X_1 \mid X_0\right)$. In other words, \textit{the prior and the target should be a pair} - there should be a clear dependence between the two.
  \item The prior should either be trivial to sample or be far cheaper to sample than the target policy. In other words, \textit{the overhead of sampling $p_{\mathcal{B}}\left(X_1 \mid X_0\right)$ must be trivial} compared to the sampling time of $X_0$. (without violating the first condition)
  \item The prior should be \textit{as close to the target distribution as possible}. (without violating the first and second conditions)
\end{itemize}

We are now ready to think about designing our priors. For the sake of comparison, we can classify the prior distributions into the following kinds:

\begin{enumerate}
  \item \textbf{Uninformative Prior}: This has no structural information about the target distribution, i.e., $p_{\mathcal{B}}\left(X_1 \mid X_0\right) = p_{\mathcal{B}}\left(X_1\right)$  
  \begin{enumerate}
    \item \textbf{Random Prior}: We use Gaussian white noise as our uninformative prior. This should reduce our framework back to DDPM. We can use this as a control to compare with the original Diffuser results.
    \end{enumerate}
  \item \textbf{Informative Prior}: This does share structural information with the target trajectory and is closer to it than random white noise.
  \begin{enumerate}
      \item \textbf{Analytical Prior}: A handcrafted or analytically calculated prior that is trivial to sample.
      \item \textbf{Learned Prior}: Something that can be cheaply sampled by an inexpensive learned model.
  \end{enumerate}
\end{enumerate}

 We also add the Number of Function Evaluations (NFE) as a variable separate from the number of diffusion steps. Since SB posterior sampling can be done in closed form (\Cref{closedformsampling}), we test out smaller NFEs while keeping the same number of diffusion steps. For DDPM, since sampling is iterative, the number of function evaluations is always equal to the number of diffusion steps.

With this architecture and theoretical foundations in place, we can move on to the experiments in the next section.
\SetPicSubDir{4-evaluation}
\SetExpSubDir{4-evaluation}

\chapter{Evaluation}
\label{ch:evaluation}

\section{Experimental Setup}

\subsection{Environment and Task}

To evaluate our model against Diffuser's default DDPM model, we design our experiment to run on the Maze2D task in the D4RL~\parencite{d4rl} suite. The task involves navigating a 2D maze by pushing a point mass to a goal location. We choose this task for a few reasons:

\begin{enumerate}
    \item The original Diffuser paper primarily compares results on the Maze2D benchmark.
    \item Since this is a maze navigation task, it is easy to construct and visualize both handcrafted and learned model priors for this task, aiding us in figuring out what's happening.
    \item The task is one of goal-conditioned planning, which gets conveniently posed as inpainting in the Diffuser architecture. Since the I\textsuperscript{2}SB paper uses image inpainting as one of the primary tasks, it makes for a much more equal comparison, without the use of any guidance functions in Diffuser.
\end{enumerate}

We use the normalized total score~\parencite{d4rl} obtained on the task over several instantiations of the task to compare performance. We compare the performance against the amount of data the two models are trained on to compare sample efficiencies. The D4RL Maze2D environment generates trajectories to train the samples on, and Diffuser uses the number of training steps to generate more or less training trajectories. We, therefore, use training steps to represent the amount of training data. 

\subsection{Task Details and Metrics}

\begin{figure}[ht!]
  \centering
  \includegraphics[width=.95\linewidth]{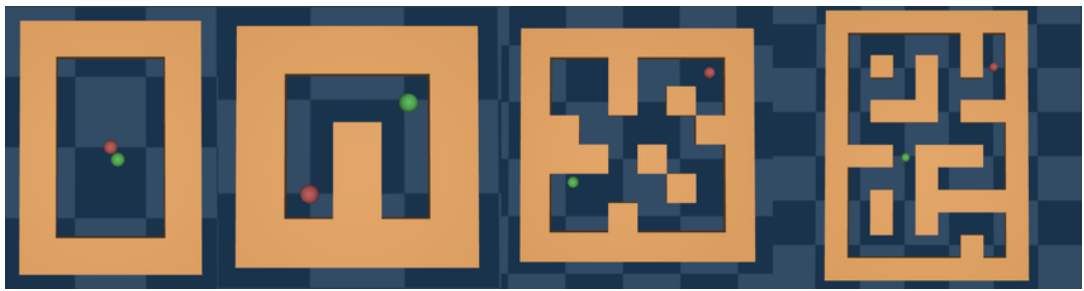}
  \vspace{\BeforeCaptionVSpace}
  \caption{The four Maze2D environments in D4RL \parencite{d4rl} - open, umaze, maze2d-medium, maze2d-large.}
  \label{maze2d_all}
\end{figure}

The environments are mazes of different sizes and complexity - open, umaze, maze2d-medium, maze2d-large. They are visualized below in \Cref{maze2d_all}.

The inputs of the task are the start location (green dot) and the goal location (red dot). The output of the planning algorithm is a trajectory - a set of state and action pairs. We sample in an open-loop fashion. The actions returned by the planner are taken in the environment one by one by a waypoint controller, and the environment returns a reward (which can be positive or negative) for each step taken. The rewards are added up and divided by an expert policy score, and the result is normalized to the 0 to 100 range. \parencite{d4rl}

\subsection{Prior Construction}

We now describe the priors constructed for our experiments and also show their trajectory on the maze2d-medium environment. In the images that follow, the paths shown are the trajectories for the pair - the prior and the target - for the same set of starting and ending points. The colors represent the progression of the trajectory - points closer in color are closer in timestep. The colorspaces for the prior and actual rollout are kept different for ease of identification. For the priors, the start positions are blue and progress in a gradient to the final position in red. For the target trajectories, we use dark blue for start positions and progress in a gradient to light blue.

\subsubsection{Analytical Prior}

For the analytical prior, we handcraft a straight-line trajectory between the start and goal locations. In other words, given a target trajectory $X_1$, our prior is another trajectory $X_0$ that joins the ends of $X_1$ by a straight line in the 2D maze environment, irrespective of the maze obstacles. This is crafted by interpolating the 2D space in the maze to determine the state space for each point, and using constant velocity in the direction of the goal in the action space. The number of timesteps are kept the same as the horizon.

This prior is trivial to evaluate, depends on the target trajectory, and has structural information about the target - it lies in the same region of the maze and connects the start and end points. It does not, however, have any information about the structure of the maze environment itself.

In \Cref{analytical} below, we show one sample of an analytical prior and its corresponding target trajectory on maze2d-medium. 

\begin{figure}[htbp!]
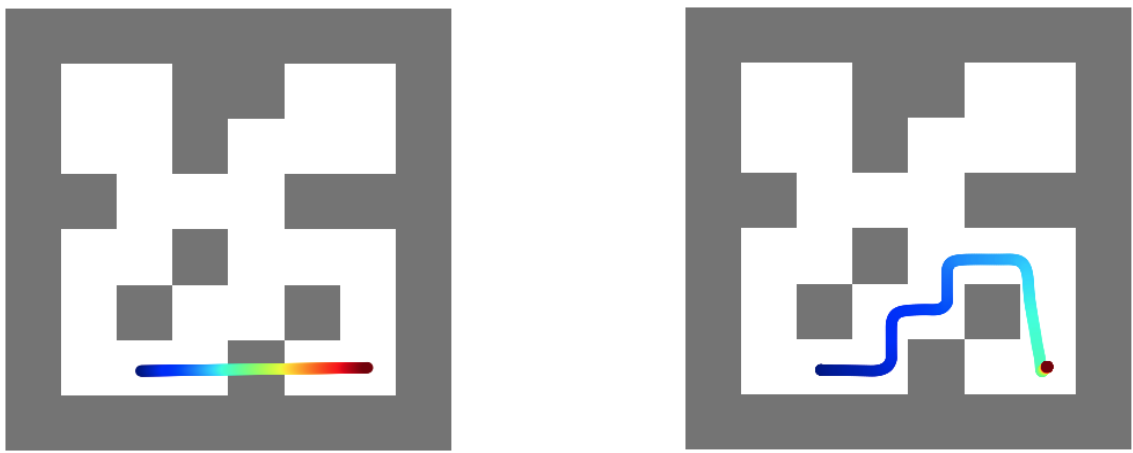

  \centering
  \begin{minipage}[b]{0.4\textwidth}
    \includegraphics[width=\textwidth]{\Pic{png}{st_line_prior}}
  \end{minipage}
  \hfill
  \begin{minipage}[b]{0.4\textwidth}
    \includegraphics[width=\textwidth]{\Pic{png}{st_line_rollout}}
  \end{minipage}
  \caption{\textbf{Left:} Analytical prior trajectory $X_0$. \textbf{Right:} Target trajectory $X_1$.}
  \label{analytical}
\end{figure}

\subsubsection{Learned Prior}

For a learned model, we want something that is inexpensive to train and sample and yet can approximate the target distribution to some degree. To that end, we train a simple model with two linear layers separated by a ReLu layer, trained to predict the prior trajectory in the same loss function and the same trajectories as the diffusion model. The code for our simple prior model in pyTorch is below - the observation dimension is the length of the states vector, while the transition dimension is the length of the joint states and actions vector:

\begin{lstlisting}[language=Python]
mlp = nn.Sequential( 
            nn.Linear(2*observation_dim, horizon*transition_dim), 
            nn.LeakyReLU(0.1), 
            nn.Linear(horizon*transition_dim, horizon*transition_dim), 
            )
\end{lstlisting}

\Cref{learned} below shows a pair of learned prior and actual target rollout on maze2d-medium. 

\begin{figure}[htbp!]
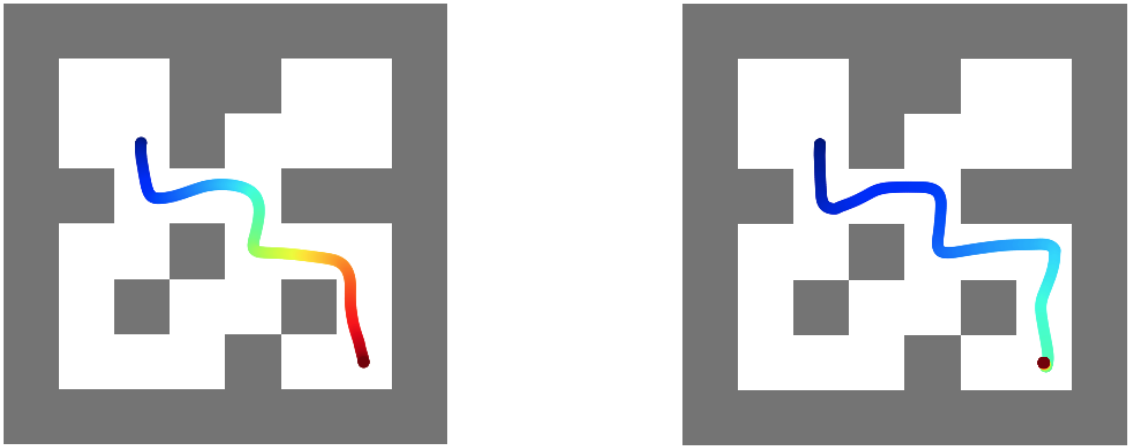

  \centering
  \begin{minipage}[b]{0.4\textwidth}
    \includegraphics[width=\textwidth]{\Pic{png}{mlp_prior}}
  \end{minipage}
  \hfill
  \begin{minipage}[b]{0.4\textwidth}
    \includegraphics[width=\textwidth]{\Pic{png}{mlp_rollout}}
  \end{minipage}
  \caption{\textbf{Left:} Learned prior trajectory $X_0$. \textbf{Right:} Target trajectory $X_1$.}
  \label{learned}
\end{figure}

The approximate trajectories not only contain information about the start and goal states, but also the structure of the maze itself, and are hence closer to the target than the analytical prior above. 

For demonstration, we chose a sample where the performance of the learned model was good enough, just to show how much information it can contain. Typically, the plans from such a simple model were much less precise and often cut through obstacles.

\subsubsection{Random Prior}

Finally, for control, we also test using Gaussian white noise as a prior - the same as what DDPM diffuses to. Each state and action value in the trajectory of the random prior is simply a random number.

\begin{figure}[htbp!]
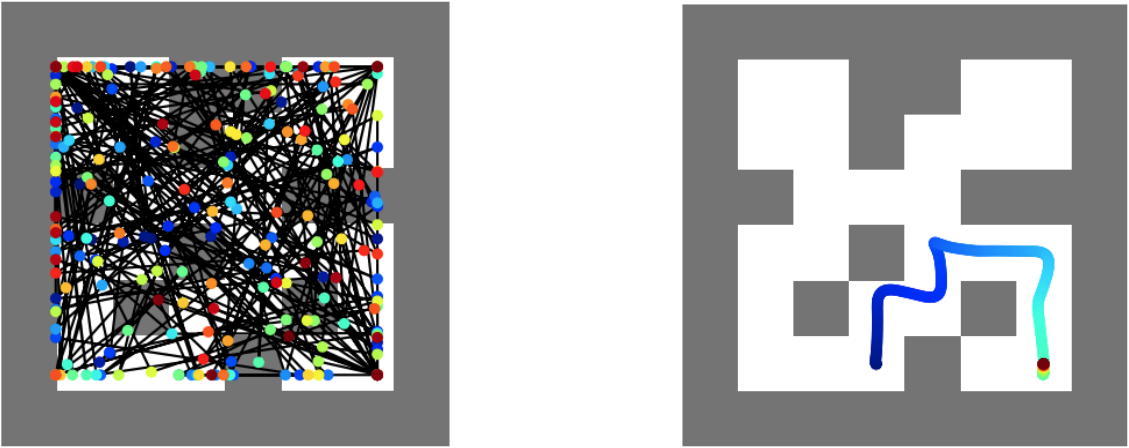

  \centering
  \begin{minipage}[b]{0.4\textwidth}
    \includegraphics[width=\textwidth]{\Pic{png}{noise_prior}}
  \end{minipage}
  \hfill
  \begin{minipage}[b]{0.4\textwidth}
    \includegraphics[width=\textwidth]{\Pic{png}{noise_rollout}}
  \end{minipage}
  \caption{\textbf{Left:} Gaussian noise prior $X_0$. \textbf{Right:} Target trajectory $X_1$.}
  \label{random}
\end{figure}

\section{Results}

In the sections below, we discuss the results of benchmarking I\textsuperscript{2}SB against DDPM on planning tasks. We ran all benchmarks for 200 episodes each, and the horizon was fixed at 256 for the maze2d-medium and 384 for the maze2d-large environment. The scores are the normalized scores as defined in D4RL~\parencite{d4rl} suite.

\subsection{Uninformative Random Prior}

\subsubsection{Diffusing from Gaussian noise - Same NFE}

\begin{figure}[ht!]
  \centering
  \includegraphics[width=.55\linewidth]{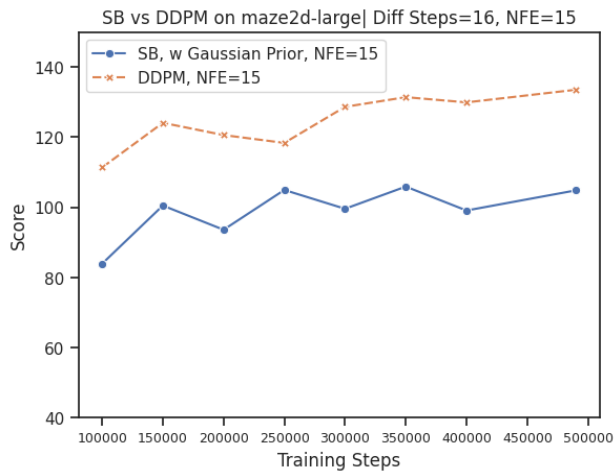}
  \caption{Schrödinger Bridge in the I\textsuperscript{2}SB formulation does not perform as well as DDPM when starting from Gaussian prior.}
  \label{score_training}
\end{figure}

Even though both of the models use the same networks and both start diffusing from Gaussian noise, we see in \Cref{score_training} a consistent drop in performance with I\textsuperscript{2}SB compared to DDPM. This suggests that when starting from pure Gaussian noise, DDPM is a better formulation than I\textsuperscript{2}SB in terms of sample quality, and since the gap increases with training samples, also a better learner. We talk about this later in \Cref{section:performance_gap}.

\subsubsection{Diffusing from Gaussian noise - across a range of NFEs}

Next, we evaluate if SB can leverage the closed-form sampling (\Cref{closedformsampling}) to outperform DDPM at lower NFEs by keeping the number of diffusion steps larger. 

From \Cref{ddpm_nfe} below, we see that DDPM struggles with NFE=1, and performance does not improve with more training. With NFE=4 and above, DDPM can reach the performance ceiling given enough data. 

\begin{figure}[htbp!]
  \centering
  \includegraphics[width=.55\linewidth]{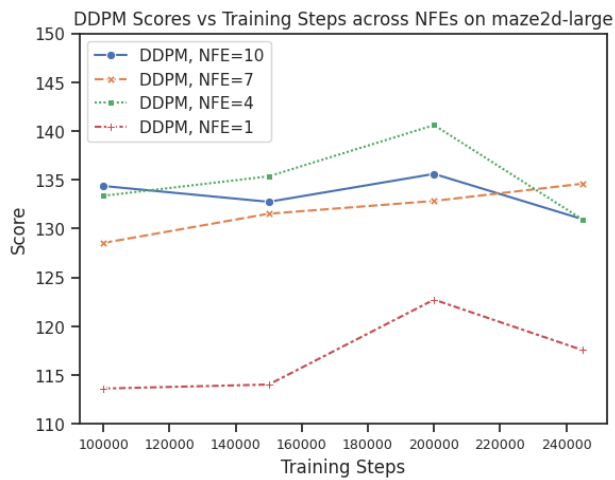}
  \vspace{\BeforeCaptionVSpace}
  \caption{Score across different diffusion steps for DDPM.}
  \label{ddpm_nfe}
\end{figure}

\begin{figure}[htbp!]
  \centering
  \includegraphics[width=.55\linewidth]{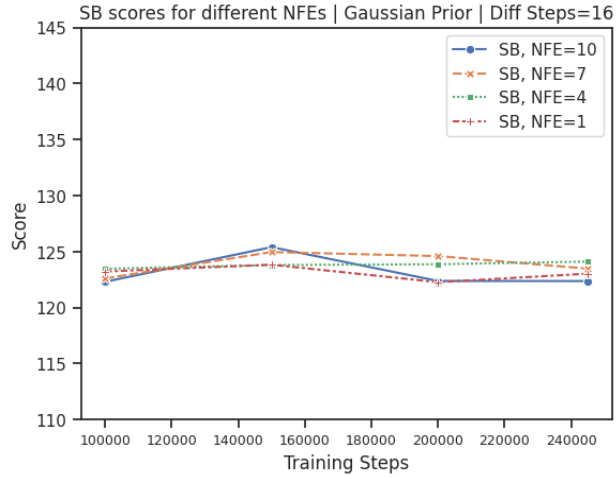}
  \vspace{\BeforeCaptionVSpace}
  \caption{Score across different NFEs for SB. Number of diffusion steps was constant at 16.}
  \label{sb_nfe}
\end{figure}

For SB (\Cref{sb_nfe}), we sample at varying NFEs while keeping the number of diffusion steps (and therefore model complexity) constant. SB performs nearly identically even at low NFEs. This suggests that closed-form sampling (\Cref{closedformsampling}) works well and prevents loss of performance at low NFEs.

\subsection{Informative Priors}

We now compare the priors we designed against each other, keeping NFE constant. We see that at the lowest training steps, more informative priors tend to perform better than less informative ones. 

\begin{figure}[ht!]
  \centering
  \includegraphics[width=.55\linewidth]{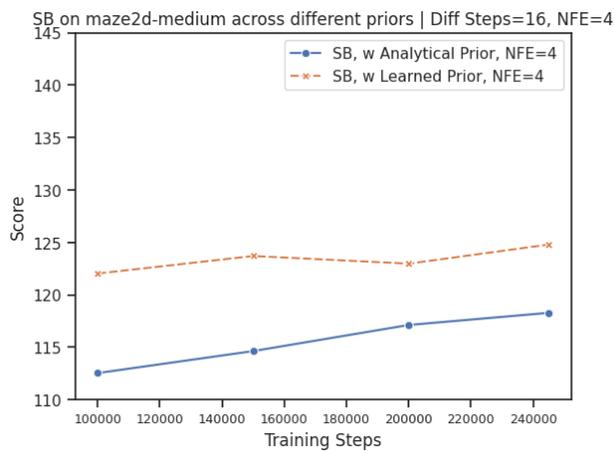}
  \vspace{\BeforeCaptionVSpace}
  \caption{Learned priors perform better than analytical priors.}
  \label{prior_score}
\end{figure}

\begin{figure}[ht!]
  \centering
  \includegraphics[width=.55\linewidth]{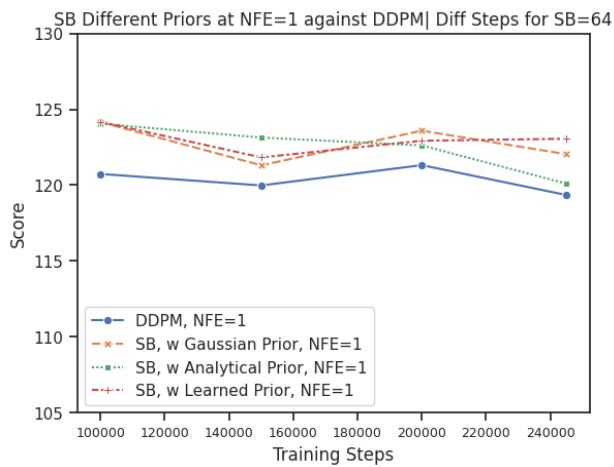}
  \vspace{\BeforeCaptionVSpace}
  \caption{I\textsuperscript{2}SB can perform better than DDPM when the NFE is low.}
  \label{sb_priors_nfe_1}
\end{figure}

In \Cref{sb_priors_nfe_1}, we compare the performance of SB models directly against DDPM at NFE=1. Even though DDPM beats I\textsuperscript{2}SB at higher model capacities, here we see that I\textsuperscript{2}SB matches or surpasses the performance of DDPM on all three priors, once again suggesting the strength of closed-form sampling. 

\section{Summary of Experimental Results}

We started by asking if starting with a prior distribution instead of Gaussian noise can help us with:
\begin{enumerate}
    \item \textit{Sample} efficiency, i.e., can we get the same performance for less training data?
    \item \textit{Sampling} efficiency, i.e., can we get the same performance with faster sampling times?
\end{enumerate}

We can now begin to answer these questions.

\subsection{Sample Efficiency}

We first note from \Cref{score_training} that given the same model complexity, the I\textsuperscript{2}SB formulation does no better than DDPM when starting from noise - in fact, it does consistently worse across the number of training samples.

However, comparing the performance at NFE=1 from \Cref{sb_priors_nfe_1}, we can see that the  I\textsuperscript{2}SB model does far better, especially at low training steps. Thus, \textit{at very low NFEs, I\textsuperscript{2}SB can maintain a much higher model complexity than DDBM, and thus have better sample efficiency}. This, of course, is a direct consequence of the way I\textsuperscript{2}SB does sampling through \Cref{closedformsampling}, which allows I\textsuperscript{2}SB to sample multiple time steps in a single function evaluation/network call.

At the same time, as the allowance for NFE increases to 4 and above, we see from \Cref{ddpm_nfe} and \Cref{prior_score} that DDPM beats I\textsuperscript{2}SB even when I\textsuperscript{2}SB uses a learned prior distribution compared to DDPM starting from pure noise. This suggests that DDPM is a better learner given enough model complexity. 

\subsection{Sampling Efficiency}

Again, it is evident from \Cref{ddpm_nfe} and \Cref{sb_nfe} that I\textsuperscript{2}SB has better sampling efficiency only at very low NFEs. As soon as we go above 4 NFE, I\textsuperscript{2}SB is not able to catch up to the sample quality of DDPM, rendering the question of sampling efficiency moot.

\subsection{On Priors}
We note from \Cref{sb_priors_nfe_1} that the more informative priors do better as the number of training steps goes up. We also note that out learned prior performs slightly better than our handcrafted analytical prior in \Cref{prior_score}.

\subsection{On the Performance Gap between I\textsuperscript{2}SB and DDPM}
\label{section:performance_gap}

A key portion of our hypothesis was whether generation can be faster if started from a closer distribution, with the implicit assumption that the generating processes that do this transport are equally good, so that a closer distribution means more efficient generation. We find that this implicit assumption did not hold in the case of  I\textsuperscript{2}SB. 

We noted in \Cref{score_training} when comparing I\textsuperscript{2}SB against DDPM, that when:

\begin{enumerate}
    \item The model complexity is kept the same for both, i.e., the same network size and the same number of diffusion steps (and therefore NFE, since DDPM needs one function evaluation per diffusion step.)
    \item And the model complexity is large enough, i.e., NFE is greater than just 1
    \item And both start from Gaussian white noise
\end{enumerate}

DDPM outperforms I\textsuperscript{2}SB irrespective of the training steps, with a consistent gap. This suggests that DDPM as an algorithm is a better learner, at least when allowed enough complexity in terms of diffusion steps.

We explain this in two ways:
\begin{enumerate}
    \item Since I\textsuperscript{2}SB is an approximation of the general Schrödinger Bridge, it makes some theoretical relaxations to permit tractability. For example, in \Cref{eq:2.13}, the authors modify the reverse drift to be the Dirac Delta distribution centered on data point $a$. They rely on the generalizing effects of neural networks to counter this effect. We also note that Diffuser's UNet, which we used for our predictions, is a smaller UNet than I\textsuperscript{2}SB's original network, because we work on much lower dimensional inputs.

    DDPM on the other hand is a more exact model without such relaxations.
    \item Some recent works have pointed out a flaw in the sampling procedure of I\textsuperscript{2}SB and other similar direct diffusion models (DDB). ~\cite{chung2023direct}, which introduces \textit{Consistent Direct Diffusion Bridges} that constantly guides the trajectory to satisfy data consistency, note the following in their paper~\parencite{chung2023direct}:

    \begin{displayquote}
        \textit{Regardless of the choice in constructing DDB, there is a crucial component that is missing from the framework. While the sampling process starts directly from the measurement (or equivalent), as the predictions $\hat{\boldsymbol{x}}_{0 \mid t}=G_\theta\left(\boldsymbol{x}_t\right)$ are imperfect and are never guaranteed to preserve the measurement condition, the trajectory can easily deviate from the desired path, while the residual blows up. Consequently, this may result in inferior sample quality, especially in terms of distortion.}
    \end{displayquote}

\end{enumerate}

The above two aspects of I\textsuperscript{2}SB are also core to its formulation. While the closed-form sampling and diffusing from a prior do lend it advantages, a poorer learning process because of the approximations, and a flawed sampling process leading to distortions over time, take away those advantages at higher model complexity. We do show (\Cref{sb_priors_nfe_1}) however that despite being a poorer learner, I2SB outperforms DDPM at very low NFE and moderate to high diffusion steps. 
\SetPicSubDir{5-discussion}

\chapter{Discussion \& Conclusion}
\label{ch:discussion}

In this thesis, we explored a novel way of planning from prior plans. We used two existing frameworks - the Diffuser way of planning using diffusion models, and the image-to-image bridged diffusion framework from I\textsuperscript{2}SB, and added our own interpretation of possible planning priors, to make the method work. Though the particular formulation of Schrödinger Bridges we used did not work as well, the overall method certainly has merit, and replacing I\textsuperscript{2}SB with bridging algorithms that are also more efficient learners might eventually yield the results we sought initially. 

Our findings may also be limited because we tried the method on one particular task and in one particular environment, however, future work on higher dimensional trajectory tasks might yield more interesting results. 

We also did not evaluate the sampling time of the priors themselves but instead assumed that operation to be trivially cheap. It could be an interesting direction to evaluate how weaker but cheaper algorithms combined with bridging methods could yield a stronger algorithm at a lower cost than using diffusion alone. 

We also did not go into the theoretical reasons for the performance of various bridging methods under various conditions as it was beyond the scope of this work, but it is a very active area of research right now. 

Better formulations of bridging models will certainly provide us with a model that efficiently generates plans from trivially learnable priors, which may significantly speed up the planning process, especially in the case of closed-loop scenarios. Aside form with readily available samples from a closer distribution, such as time series forecasting, and modelling temporal data such as video generation, could also benefit from such advances.

\bookmarksetup{startatroot}
\printbibliography[heading=bibintoc]

\end{document}